\documentclass[conference]{IEEEtran}
\IEEEoverridecommandlockouts
\usepackage{cite}
\usepackage{amsmath,amssymb,amsfonts}
\usepackage{algorithmic}
\usepackage{graphicx}
\usepackage{textcomp}
\usepackage{xcolor}
\def\BibTeX{{\rm B\kern-.05em{\sc i\kern-.025em b}\kern-.08em
    T\kern-.1667em\lower.7ex\hbox{E}\kern-.125emX}}

\begin{document}
\title{AppendiGrade: An XAI-Enhanced Deep Learning Framework for Grading Appendicitis in Ultrasound with Gaussian Blur and Grad-CAM}
\author{
\IEEEauthorblockN{Fahad Ahammed}
\IEEEauthorblockA{
\textit{Computer Science and}
\textit{Information Technology}\\
\textit{Ca' Foscari University of Venice}\\
Venice, Italy\\
fahadahbd@gmail.com
}
\and
\IEEEauthorblockN{Omar Faruq Shikdar}
\IEEEauthorblockA{\textit{Computer Science and Engineering} \\
\textit{East West University}\\
Dhaka, Bangladesh \\
omorfaruk549@gmail.com}
\and
\IEEEauthorblockN{Navid Zaman}
\IEEEauthorblockA{
\textit{Department of Digital Business}\\
\textit{and Innovation}\\
\textit{Tokyo International University}\\
Tokyo, Japan\\
navidzaman.xyz@gmail.com
}
\and
\IEEEauthorblockN{Md Tahsin}
\IEEEauthorblockA{\textit{Computer Science and Engineering} \\
\textit{East West University}\\
Dhaka, Bangladesh \\
tahsin30899@gmail.com}
\and
\IEEEauthorblockN{Md. Nawab Yousuf Ali}
\IEEEauthorblockA{\textit{Computer Science and Engineering} \\
\textit{East West University}\\
Dhaka, Bangladesh \\
nawab@ewubd.edu}
\and
\IEEEauthorblockN{Golam Sorwar}
\IEEEauthorblockA{\textit{Computer Science and Engineering} \\
\textit{Southern Cross University}\\
Lismore, Australia \\
golam.sorwar@scu.edu.au}
}
\maketitle

\begin{abstract}
Appendicitis is one of the most common abdominal emergencies worldwide and requires prompt diagnosis and treatment to prevent life-threatening conditions. However, accurately differentiating complicated cases, such as perforation or abscess formation from uncomplicated appendicitis remains a significant clinical challenge. Among other methods, ultrasound is a safer and more cost-efficient diagnostic technique because of a lack of radiation exposure. In this research, an advanced system capable of automatically detecting complicated appendicitis from ultrasound images was developed. A dataset consisting of 4679 ultrasound images with 5 classes perforated, abscess, acute, appendicolith, and normal was used for the proposed model training and testing. Four pretrained deep learning models DenseNet201, InceptionV3, ConvNextTiny, and VGG19, have been employed for detecting and classifying complicated appendicitis. In the initial configuration, InceptionV3 achieved the second highest accuracy, with a value of 69.21\%. Owing to suboptimal performance with raw images, further optimization techniques, including image preprocessing, hyperparameter tuning, model fine-tuning, and image sharpening, were applied. These enhancements significantly improved the model’s performance, with an accuracy of 95.58\% for InceptionV3. The model performance is then explained with gradient-weighted class activation mapping (Grad-CAM), which creates a heatmap of the regions responsible for the model’s prediction of the infected areas. This could make crosschecking with experts much easier. 
\end{abstract}


\section{Introduction}
One of the most common abdominal emergencies in the world is appendicitis [1]. From 1990-2019, the incidence of appendicitis increased by 63.55\% worldwide [2]. Early and accurate diagnosis is critical for timely treatment, but traditional imaging methods are time-consuming, require expert radiologists, and involve significant training and cost [3]. There are mainly four types of complicated appendicitis cases that can occur. Acute appendicitis is sudden and severe inflammation of the appendix [4]. Perforated appendicitis occurs when infection causes a hole, allowing spread to the abdomen [5]. Appendicitis abscess involves pockets of infection formed as a complication [6]. Appendicolith is a calcified deposit or stone-like object that forms within the appendix [7].

This study aims to address previous gaps by developing a deep learning model for detecting different types of complicated appendicitis from a comprehensive dataset of 4,679 ultrasound images across five classes (acute, perforated, abscess, appendicolith, and normal). Four pretrained models (DenseNet201, InceptionV3, ConvNextTiny, and VGG19) were employed. The models were initially trained on raw images, followed by extensive image preprocessing (Gaussian blur and unsharp masking), layer freezing, and hyperparameter tuning. Grad-CAM was integrated to generate heatmaps highlighting affected areas.

The remainder of this paper is structured as follows. Section 2 discusses the previous study. Section 3 presents the dataset description, preprocessing methodology, and model implementation results. Section 4 compares results across models and before and after optimization. Section 5 concludes with remarks and future work.

\section{literature review}
Artificial intelligence-based automated diagnostic systems can provide an efficient solution by reducing reliance on radiologists while offering instant results. Most current research focuses primarily on CT and MRI, with several limitations. Byun achieved 85.5\% accuracy but was limited to confirmed cases and CT imaging [8]. Walid achieved an AUC of 0.868 but required extensive training data [9]. Rajpurkar introduced AppendiXNet with 72.5\% accuracy, though the dataset was too small [10]. Park validated a CNN model with CT data but was limited by a small dataset [11]. Mijwil tested multiple models, achieving 83.75\% accuracy but faced interpretability challenges [12]. Pati achieved 92.15\% accuracy with AdaBoost but lacked model interpretability [13]. Akbulut employed Catboost with 88.2\% accuracy but was limited by retrospective data [14]. Marcinkevičs achieved 81.9\% accuracy but faced potential sampling bias [15]. Akmeşe achieved 95.31\% accuracy but was limited to medical records [16]. Although research has used CT and MRI, most studies have been conducted on small datasets, and no research has integrated explainable AI for model interpretability. Ultrasound is the safest option because it does not involve radiation exposure. Poortman compared logistic regression and random forest with CT and sonography images, achieving 83\% and 78\% accuracy, but lacked advanced ML models [17]. Alelyani achieved 85.7\% sensitivity with ultrasound but faced challenges with patient variability [18]. Erdas deployed five pretrained models, with EfficientNetV2 achieving 96\% accuracy, but it can detect only acute appendicitis and does not account for other types [19].

On the basis of the above discussion, existing studies using ultrasound for appendicitis diagnosis are limited by small datasets, a lack of explainable AI, and a focus on binary or at most three-class classification, leaving a gap in detecting all major forms of complicated appendicitis.

\section{Research Methodology}
A dataset of four types of complicated appendicitis was collected. The models were first trained using raw images, but the results were not good enough. Therefore, image preprocessing, layer freezing, and hyperparameter tuning were applied to improve performance. Grad-CAM was then used to explain the model’s decisions

\subsection{Dataset Description}

The dataset encompasses 4679 prelabelled ultrasound images from Roboflow [20], categorized into five classes: abscess, appendicolith, perforated, acute, and normal. A certified physician reviewed and cross-checked the annotations to ensure label accuracy. This dataset is unique in its size, diversity, and balanced distribution across all five classes, providing a valuable opportunity to analyze four distinct types of complicated appendicitis, representing a significant advancement over existing datasets

\begin{figure}[htbp]
\centerline{\includegraphics[width=0.4\textwidth]{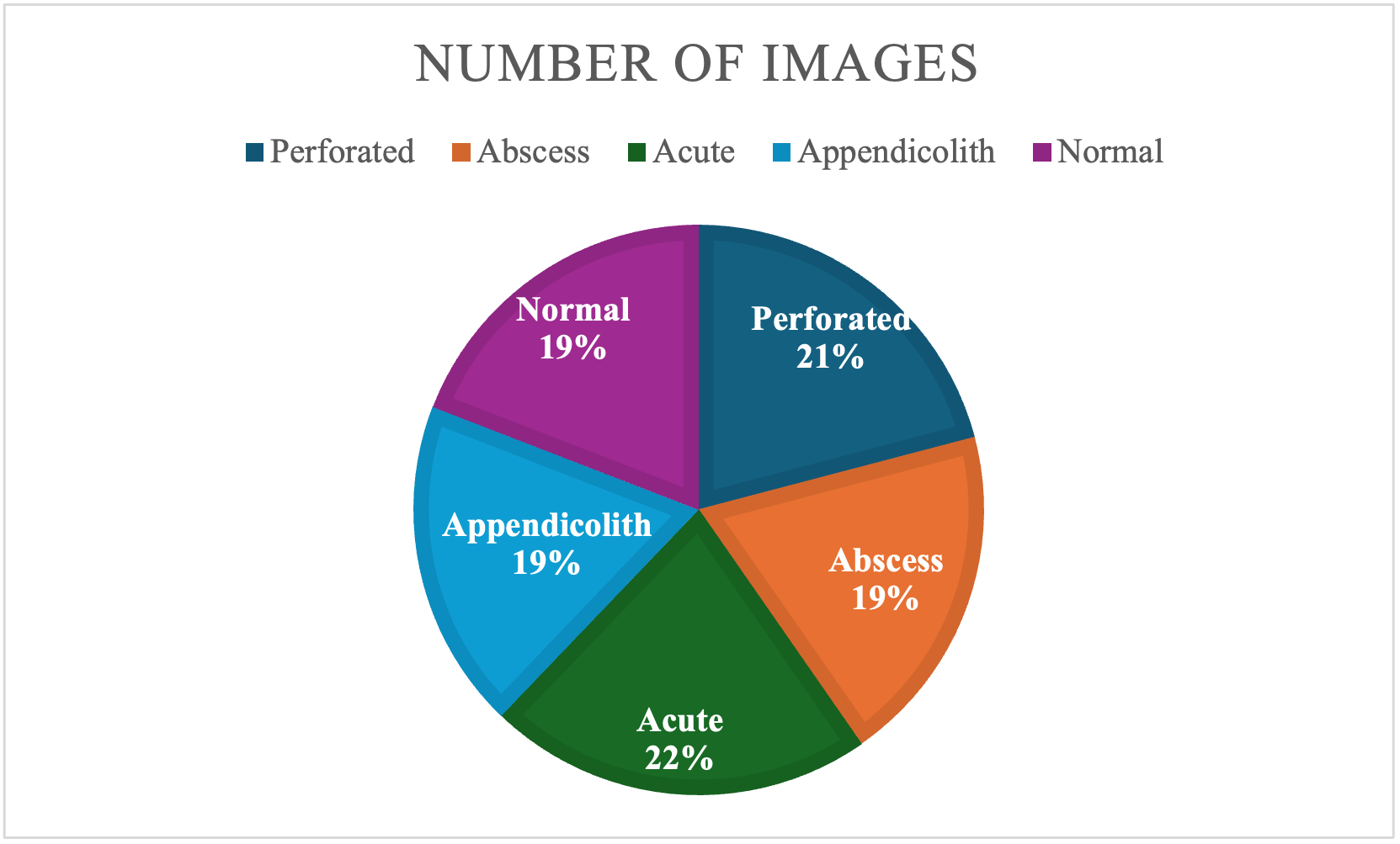}}
\caption{Sample distribution of images in each class.}
\label{fig}
\end{figure}

\subsection{Dataset Pre-processing}

Initially, all images were resized to $200*200$ pixels to eliminate unnecessary details. During normalization, pixel values were scaled to facilitate faster convergence. Augmentation techniques, such as rotation and flipping, were optionally applied to increase dataset diversity. The model trained on raw images performed poorly due to speckle noise and poor tissue contrast. To enhance performance, Gaussian blur was first applied to reduce noise, followed by an unsharp mask to sharpen images. This combination improves edge detection by reducing noise and enhancing contrast. If Gaussian blur is applied to an original image \textit{I(x,y)} to create a smoothed version, denoted as \textit{I\_smooth(x,y)}, it can be expressed as:
\begin{equation}
    I\_smooth(x,y) = (I * G\_\sigma)(x,y)
\end{equation}
\begin{equation}
    G\_\sigma(x,y)=\frac{1}{2\pi\sigma^2}*e^{-\frac{1}{2}\frac{(x-x_n)^2}{t}}
\end{equation}

The smoothed images are generated by convolving \textit{I(x,y)} with a Gaussian kernel. The smoothed images are subtracted from the original to create a mask highlighting edges:
\begin{equation}
    Mask(x,y) = I(x,y) - I\_smooth(x,y)
\end{equation}
Finally, the mask is added to the original image to obtain the sharpened image:
\begin{equation}
    I\_sharp(x, y) = I(x, y) + amount * Mask(x, y)
\end{equation}

\begin{figure}[htbp]
\centerline{\includegraphics[width=0.4\textwidth]{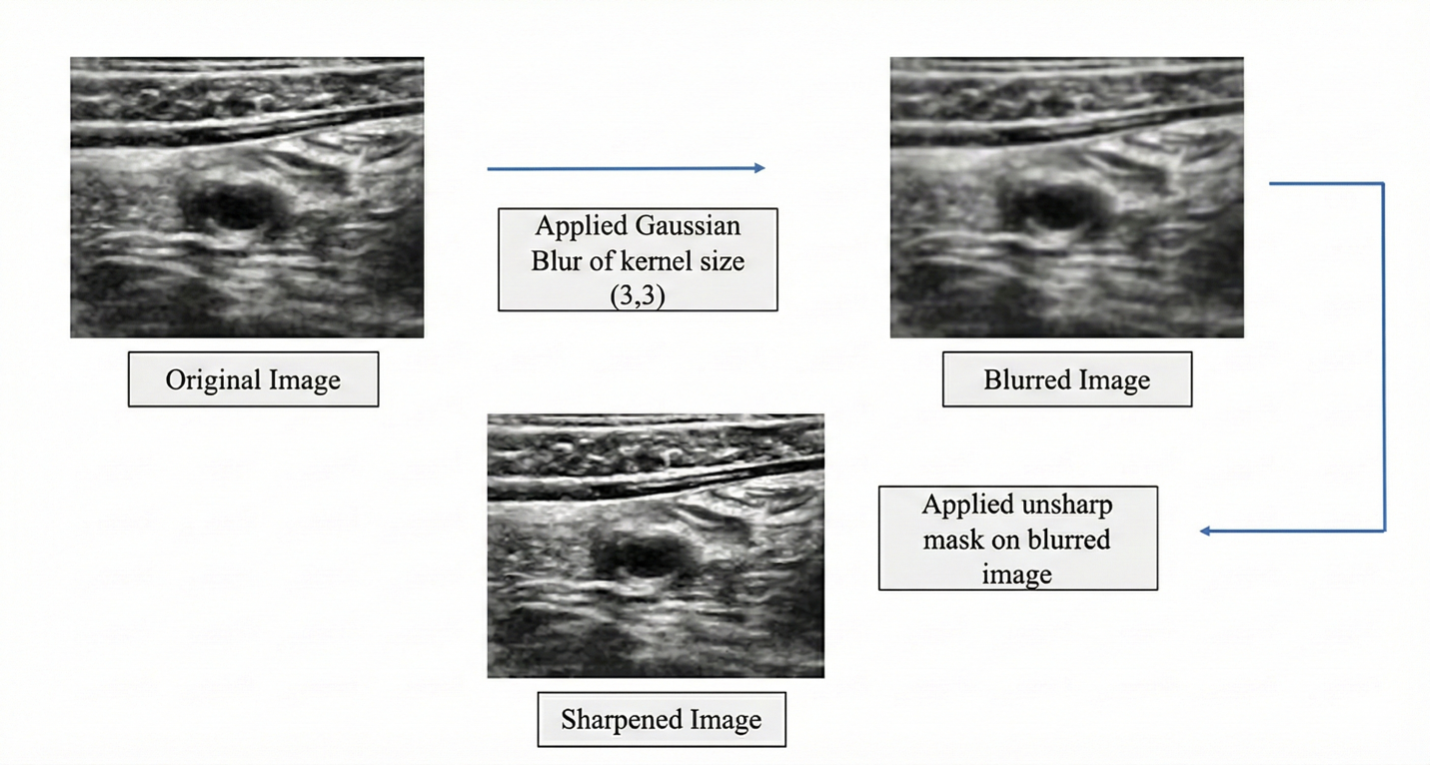}}
\caption{Image Sharpening via Gaussian blur and an unsharp mask.}
\label{fig}
\end{figure}

To achieve a comprehensive evaluation of the models, the dataset was split into 70\% for training, 15\% for validation, and 15\% for testing via stratified sampling. This splitting strategy is standard practice when deep learning models are used [21]. The benefit of stratified sampling is that the class distribution remains the same across all newly created subsets after splitting.
\subsection{Pretrained models}
Four pretrained models were employed: DenseNet201, InceptionV3, ConvNextTiny, and VGG19. DenseNet201's hierarchical feature extraction is helpful for complex ultrasound images, while InceptionV3's multiscale features are effective for multiclass classification. ConvNextTiny, a lightweight model, was selected to assess resource-efficient performance, and VGG19 was chosen as a benchmark for its simple architecture.

\subsection{Model fine tuning and hyperparameter optimization}
Each deep learning model had specific layers frozen during fine-tuning: 469 of 706 layers for DenseNet201 and 139 of 310 for InceptionV3. Freezing combinations were tested via grid search. Hyperparameters (learning rate, patience, epochs, batch size, optimizer) were optimized via random search. The Adam optimizer was used. Fig. 3 is a parallel coordinate plot showing relationships between L2 regularization, learning rate, and performance metrics to identify optimal hyperparameter combinations for better generalization.

\begin{figure}[htbp]
\centerline{\includegraphics[width=0.45\textwidth]{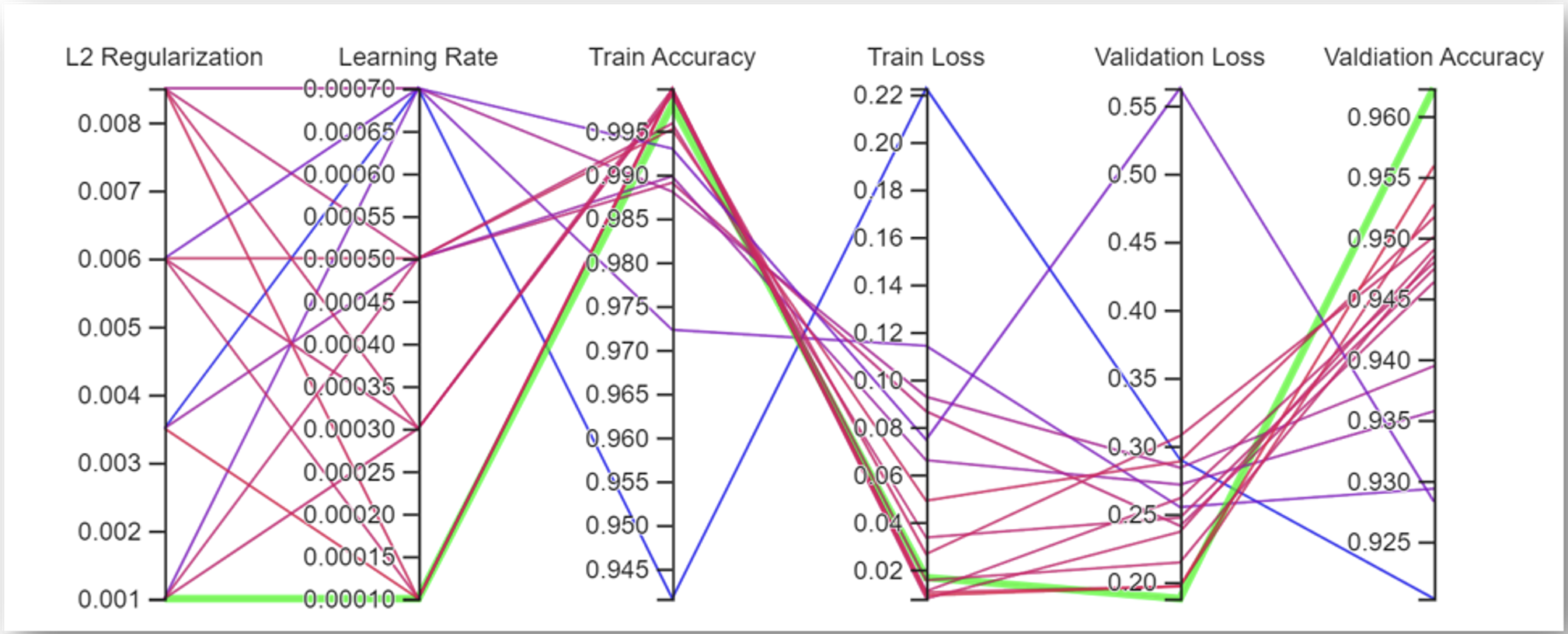}}
\caption{Hyperparameter tuning combination of DenseNet201}
\label{fig}
\end{figure}

Table 1 on the other hand presents the results of hyperparameter tuning iterations for the Inception V3 model. Both illustrate variations in the learning rate and L2 regularization, along with the corresponding validation loss for each iteration. The primary objective was to identify the optimal combination of hyperparameters that minimizes validation loss, thus improving overall model performance.

\begin{table}[htbp]
\centering
\caption{Hyperparameter tuning iteration for the InceptionV3 Model}
\label{tab:inceptionv3_hyperparameter}
\begin{tabular}{|c|c|c|c|}
\hline
\textbf{Iteration} & \textbf{Learning Rate} & \textbf{L2 Regularization} & \textbf{Validation loss} \\ 
\hline
00 & 0.0007 & 0.0035000 & 0.28976 \\
01 & 0.0007 & 0.0010000 & 0.25492 \\
02 & 0.0001 & 0.0010000 & 0.18773 \\
03 & 0.0005 & 0.0085000 & 0.24111 \\
04 & 0.0005 & 0.0035000 & 0.27150 \\
05 & 0.0007 & 0.0085000 & 0.28386 \\
06 & 0.0005 & 0.0060000 & 0.28386 \\
07 & 0.0001 & 0.0035000 & 0.19674 \\
08 & 0.0005 & 0.0010000 & 0.24710 \\
09 & 0.0001 & 0.0085000 & 0.19684 \\
10 & 0.0003 & 0.0085000 & 0.30720 \\
11 & 0.0007 & 0.0060000 & 0.56261 \\
12 & 0.0001 & 0.0060000 & 0.23685 \\
13 & 0.0003 & 0.0060000 & 0.21428 \\
14 & 0.0003 & 0.0010000 & 0.26137 \\
\hline
\end{tabular}
\end{table}

\subsection{Gradient-weighted class activation mapping}
In this study, explainable AI (XAI) Fig. 4 is used to enhance model interpretability using Grad-CAM, which highlights regions of interest in medical images and provides insights into which features influenced the model's predictions for different types of appendicitis. This approach improved transparency and facilitated understanding of the factors contributing to its decisions. For a given input image, I, and a class score $y^c$ for class c, the gradient of the class score with respect to the feature maps $A^k$ of a specific layer $k$ is computed as: $\frac{\partial y^{c}}{\partial A^{k}}$.
The global average of the gradients across all the spatial locations is
\begin{equation}
    \alpha_{c}^{k}=\frac{1}{Z}\sum_{i,j}\frac{\partial y_{c}}{\partial A_{ij}^{k}}
\end{equation}
Class activation map
\begin{equation}
    L_{\text{Grad-CAM}}=\text{ReLU}\left(\sum_{k}\alpha_{c}^{k}A^{k}\right)
\end{equation}
\begin{figure}[htbp]
\centerline{\includegraphics[width=0.45\textwidth]{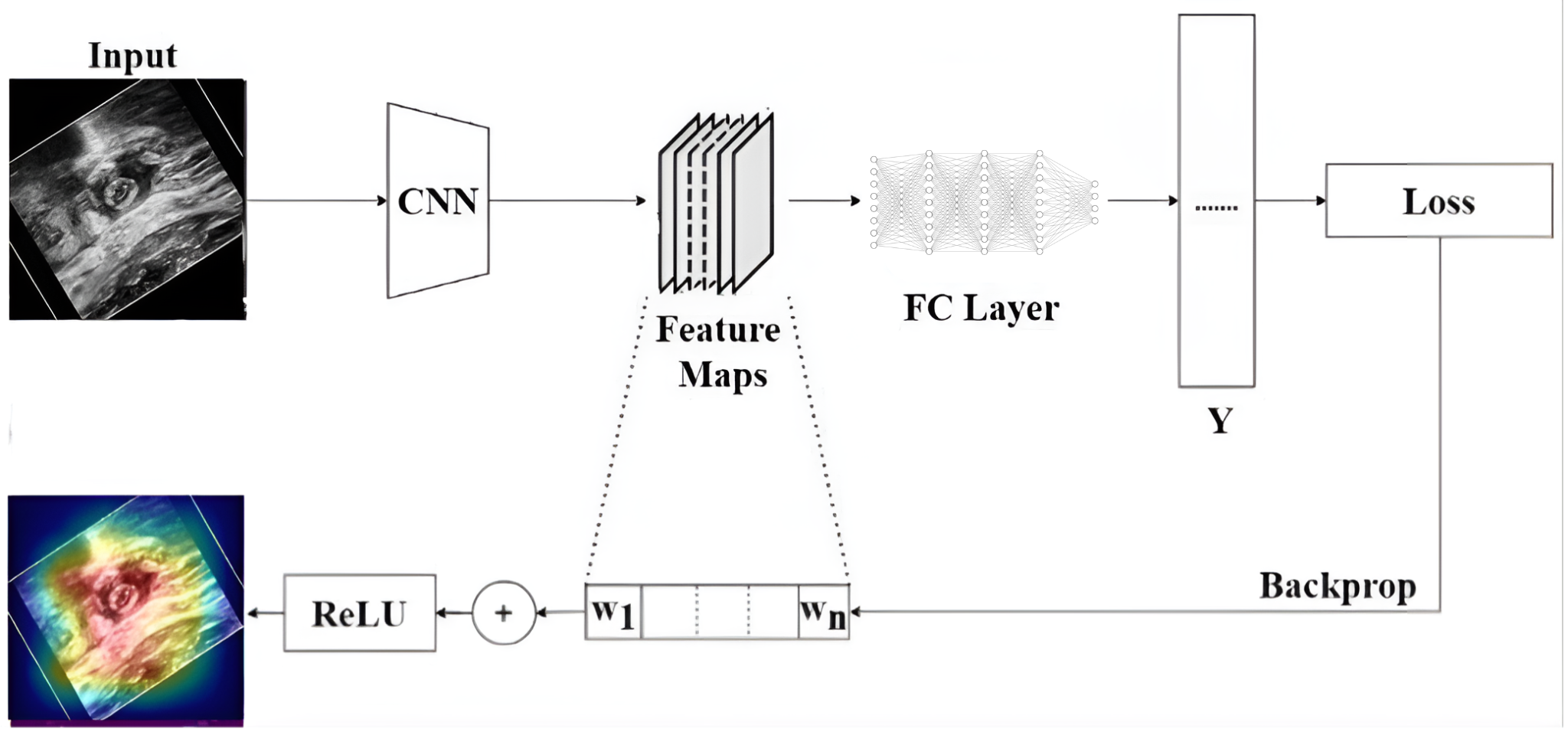}}
\caption{Explainable AI (XAI) technique Gradient-weighted Class Activation Mapping (Grad-CAM)}
\label{fig}
\end{figure}

\section{Experimental results}
Table 2 presents the performance results of each model before optimization when the models were trained with only raw images.
\begin{table}[htbp]
\centering
\caption{Results comparison of different models before optimization}
\label{tab:model_comparison}
\begin{tabular}{|c|c|c|c|c|}
\hline
\textbf{Model} & \textbf{Precision} & \textbf{F1-Score} & \textbf{Recall} & \textbf{Accuracy} \\
\hline
DenseNet201 & 0.7248 & 0.7169 & 0.7208 & 0.7208 \\
InceptionV3 & 0.6988 & 0.6892 & 0.6921 & 0.6921 \\
ConvNeXtTiny & 0.6722 & 0.6031 & 0.6103 & 0.6103 \\
VGG19 & 0.5159 & 0.2851 & 0.3586 & 0.3586 \\
\hline
\end{tabular}
\end{table}
Among the four models, DenseNet201 achieved the highest performance, with an accuracy of 72.08\%, closely followed by InceptionV3 at 69.21\%. Their precision, recall, and F1 scores were 72.48\%, 72.08\%, 71.69\% and 69.88\%, 69.21\%, 68.92\%, respectively. ConvNextTiny achieved 61.03\% accuracy, while VGG19 performed the worst at 35.86\%. Due to these suboptimal results, we applied image sharpening, fine-tuning, and hyperparameter tuning. These optimization techniques were applied only to the two best models to reduce computational cost. After optimization, DenseNet201 accuracy increased from 72\% to 94\%, with precision, recall, and F1 score all exceeding 90\% presented in Table 3.
\begin{table}[htbp]
\centering
\caption{Classification Report of DenseNet201 after Optimization}
\label{tab:densenet_classification}
\begin{tabular}{|c|c|c|c|c|}
\hline
 & \textbf{Precision} & \textbf{Recall} & \textbf{F1-score} & \textbf{Support} \\
\hline
Abscess & 0.96 & 0.98 & 0.97 & 665 \\
Acute & 0.90 & 0.94 & 0.92 & 657 \\
Appendicolith & 0.92 & 0.97 & 0.94 & 569 \\
Normal & 0.97 & 0.87 & 0.92 & 675 \\
Perforated & 0.95 & 0.95 & 0.95 & 654 \\
\hline
accuracy & 0.94 & 0.94 & 0.94 & 3220 \\
macro average & 0.94 & 0.94 & 0.94 & 3220 \\
\hline
weighted average & \multicolumn{3}{c|}{0.94} & 3220 \\
\hline
\end{tabular}
\end{table}

Similarly, Table 4 presents how InceptionV3's performance improved after optimization, with the improvement being even greater than that of DenseNet201. Its accuracy increased from 69\% to 96\%. Accuracy was not the only metric that improved, precision, recall, and F1 score all rose above 90\%, increasing from approximately 69\%.  
\begin{table}[htbp]
\centering
\caption{Classification Report of InceptionV3 after optimization}
\label{tab:inception_classification}
\begin{tabular}{|c|c|c|c|c|}
\hline
 & \textbf{Precision} & \textbf{Recall} & \textbf{F1 score} & \textbf{Support} \\
\hline
Abscess & 0.96 & 0.98 & 0.97 & 332 \\
Acute & 0.94 & 0.92 & 0.93 & 328 \\
Appendicolith & 0.98 & 0.97 & 0.97 & 284 \\
Normal & 0.96 & 0.96 & 0.96 & 337 \\
Perforated & 0.94 & 0.96 & 0.95 & 327 \\
\hline
accuracy & 0.96 & 0.96 & 0.96 & 1608 \\
macro average & 0.96 & 0.96 & 0.96 & 1608 \\
\hline
weighted average & \multicolumn{3}{c|}{0.96} & 1608 \\
\hline
\end{tabular}
\end{table}
 
Fig. 5 shows each model's accuracy progression over training epochs. While both DenseNet201 and InceptionV3 achieved high training accuracy, InceptionV3 demonstrated more stable and consistent performance. DenseNet201 exhibited significant fluctuations in validation accuracy and loss, indicating potential overfitting and instability. In contrast, InceptionV3 displayed smoother convergence with less unpredictability in validation metrics, suggesting better generalizability and reliability with fewer signs of overfitting. Table 5 compares model performance, showing InceptionV3 achieved 99.38\% training accuracy and 95.58\% test accuracy, while DenseNet201 achieved 99.47\% training and 93.94\% test accuracy.

\begin{table}[htbp]
\centering
\caption{Comparison of the accuracy of the InceptionV3 and DenseNet201 models after optimization}
\label{tab:accuracy_comparison}
\begin{tabular}{|c|c|c|}
\hline
\textbf{Model} & \textbf{Train Accuracy} & \textbf{Test Accuracy} \\
\hline
DenseNet201 & 99.47\% & 93.94\% \\
InceptionV3 & 99.38\% & 95.58\% \\
\hline
\end{tabular}
\end{table}

InceptionV3's superior test accuracy indicates better generalizability. This is attributed to its multiscale feature extraction, which handles complex ultrasound textures effectively, and its fewer parameters, enabling better generalization on unseen data. Additionally, InceptionV3's architecture is inherently better at capturing features from sharpened, noise-reduced images from the comparison, we can clearly conclude that InceptionV3 is the best model. Now, let us examine how optimization improved the model's performance by comparing its performance before and after optimization. The impact of optimization on the performance of InceptionV3 is evident in the training and validation loss and accuracy curves Figs. 6-7.

\begin{figure}[htbp]
\centerline{\includegraphics[width=0.4\textwidth]{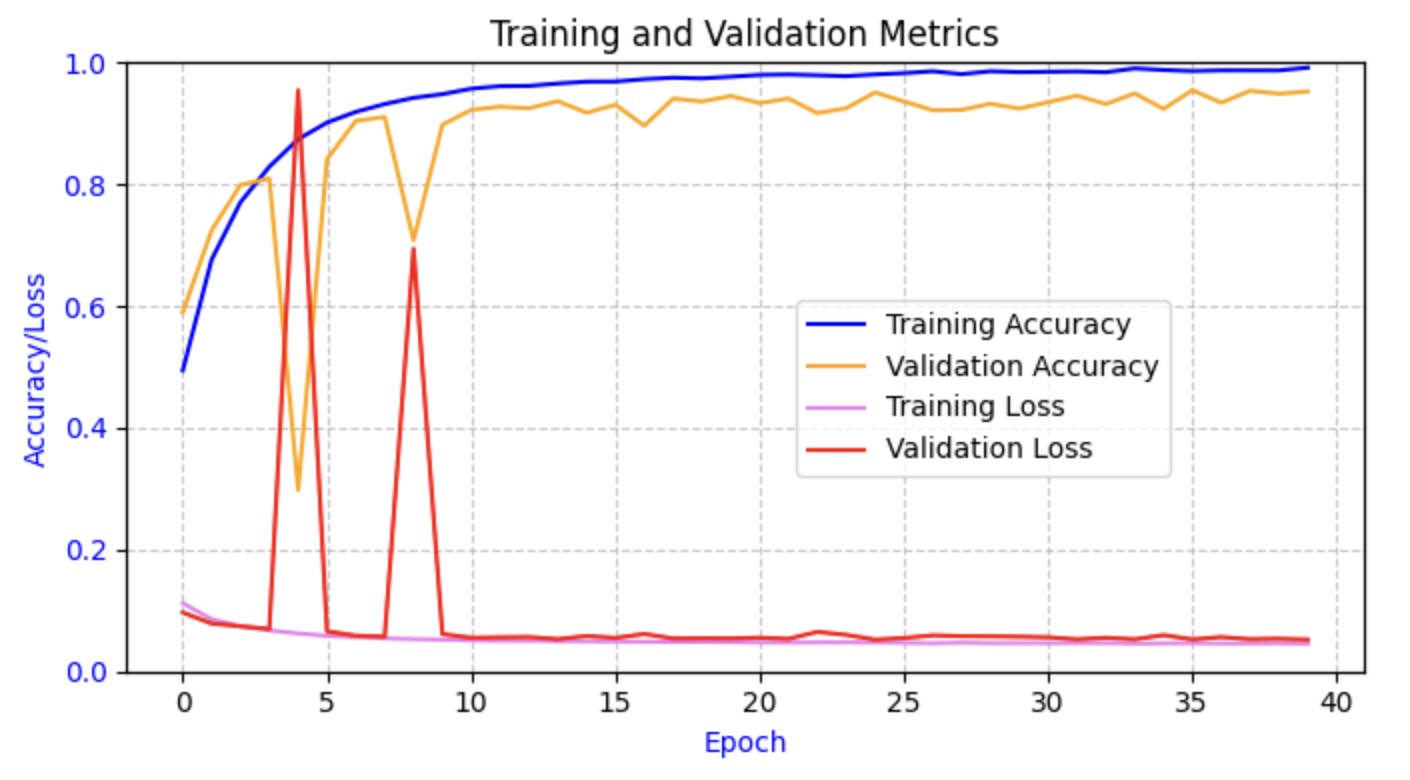}}
\centerline{\includegraphics[width=0.4\textwidth]{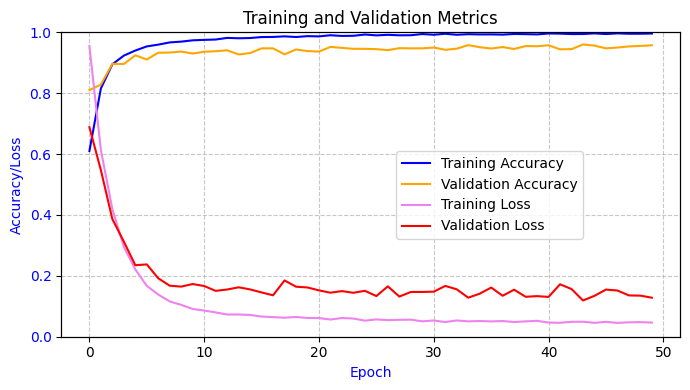}}
\caption{DenseNet201 and InceptionV3 models training curves after optimization.}
\label{fig}
\end{figure}

\begin{figure}[htbp]
\centerline{\includegraphics[width=0.4\textwidth]{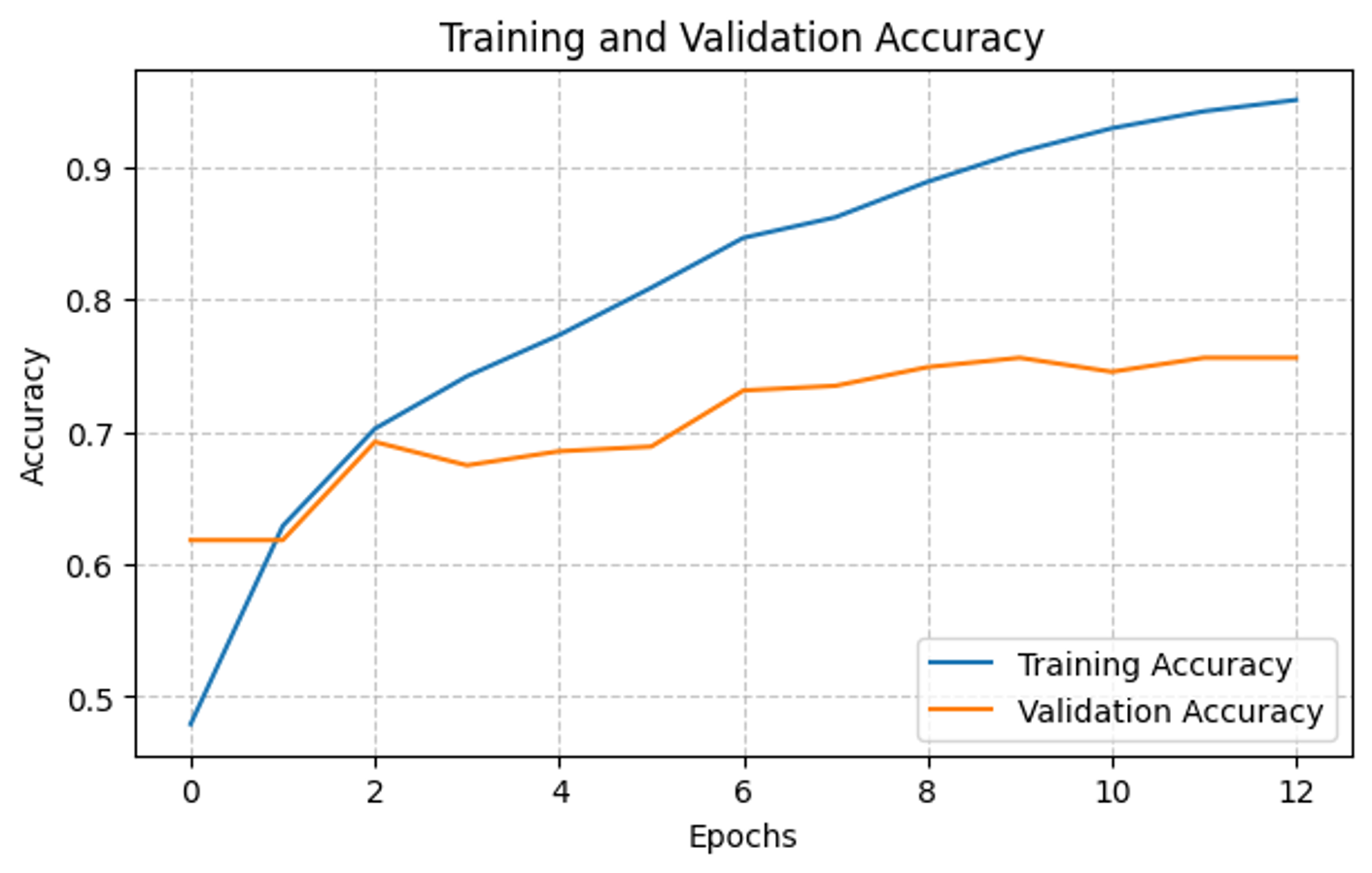}}
\centerline{\includegraphics[width=0.4\textwidth]{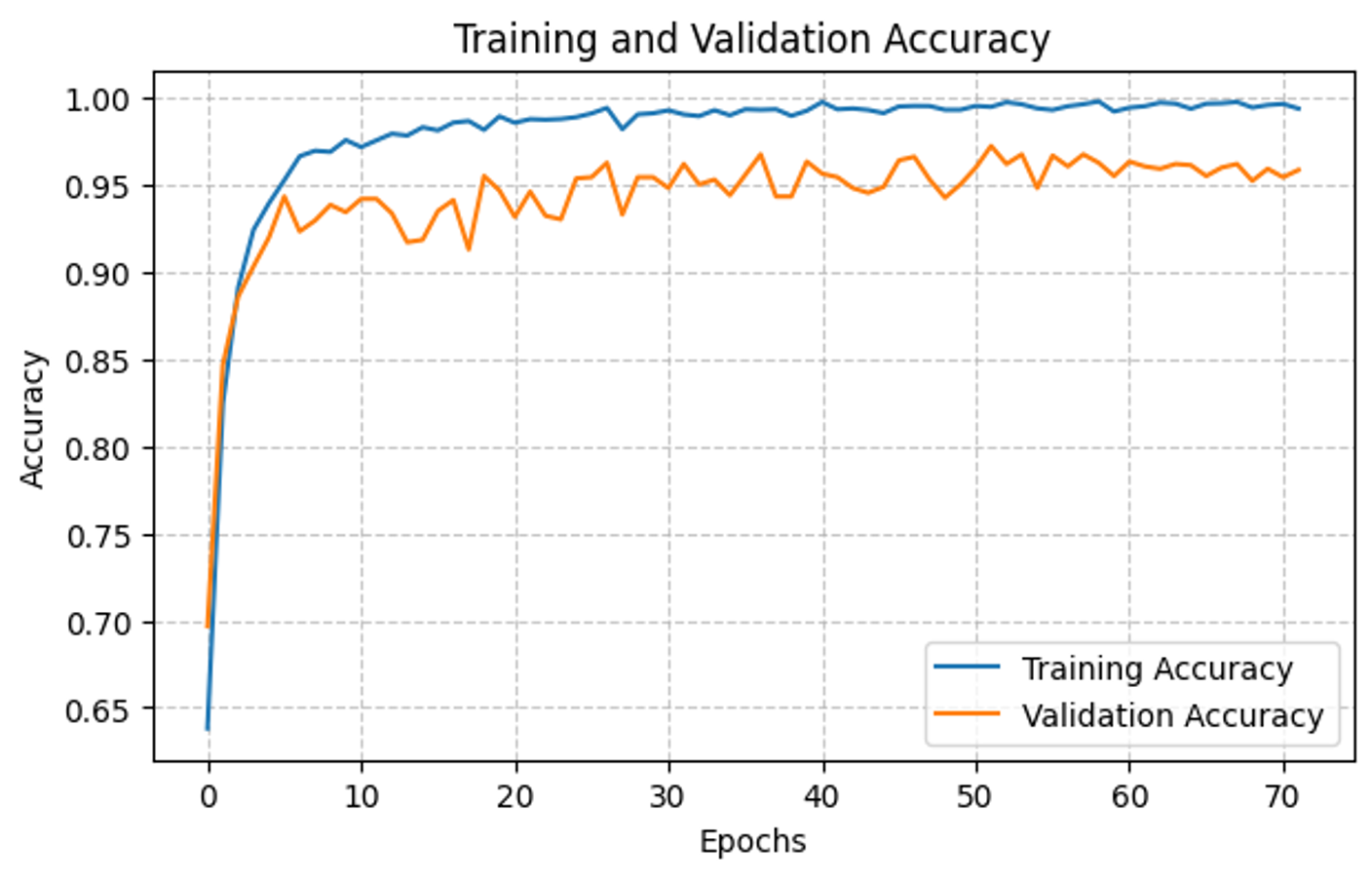}}
\caption{Accuracy comparison of InceptionV3 before and after optimization}
\label{fig}
\end{figure}

Before optimization, a clear disparity in performance was evident, with low validation accuracy and high validation loss compared with training metrics. After optimization, the difference became negligible, with training and validation accuracy and loss closely aligned. The comparative analysis in Fig. 8 highlights a significant improvement in InceptionV3 performance across all classes after training with enhanced images. For the abscess class, precision, recall, and F1 score increased from below 90\% (precision 61\%) to 96\%, 98\%, and 97\%, respectively. Similar improvements were observed for the other classes. For the acute class, precision, recall, and F1 score improved from 71\%, 59\%, and 65\% to 94\%, 92\%, and 93\%, respectively. The most dramatic change was in the appendicolith class, where metrics increased from 65\%, 55\%, and 60\% to 98\%, 97\%, and 97\%, representing a 30-37\% improvement.

\begin{figure}[htbp]
\centerline{\includegraphics[width=0.4\textwidth]{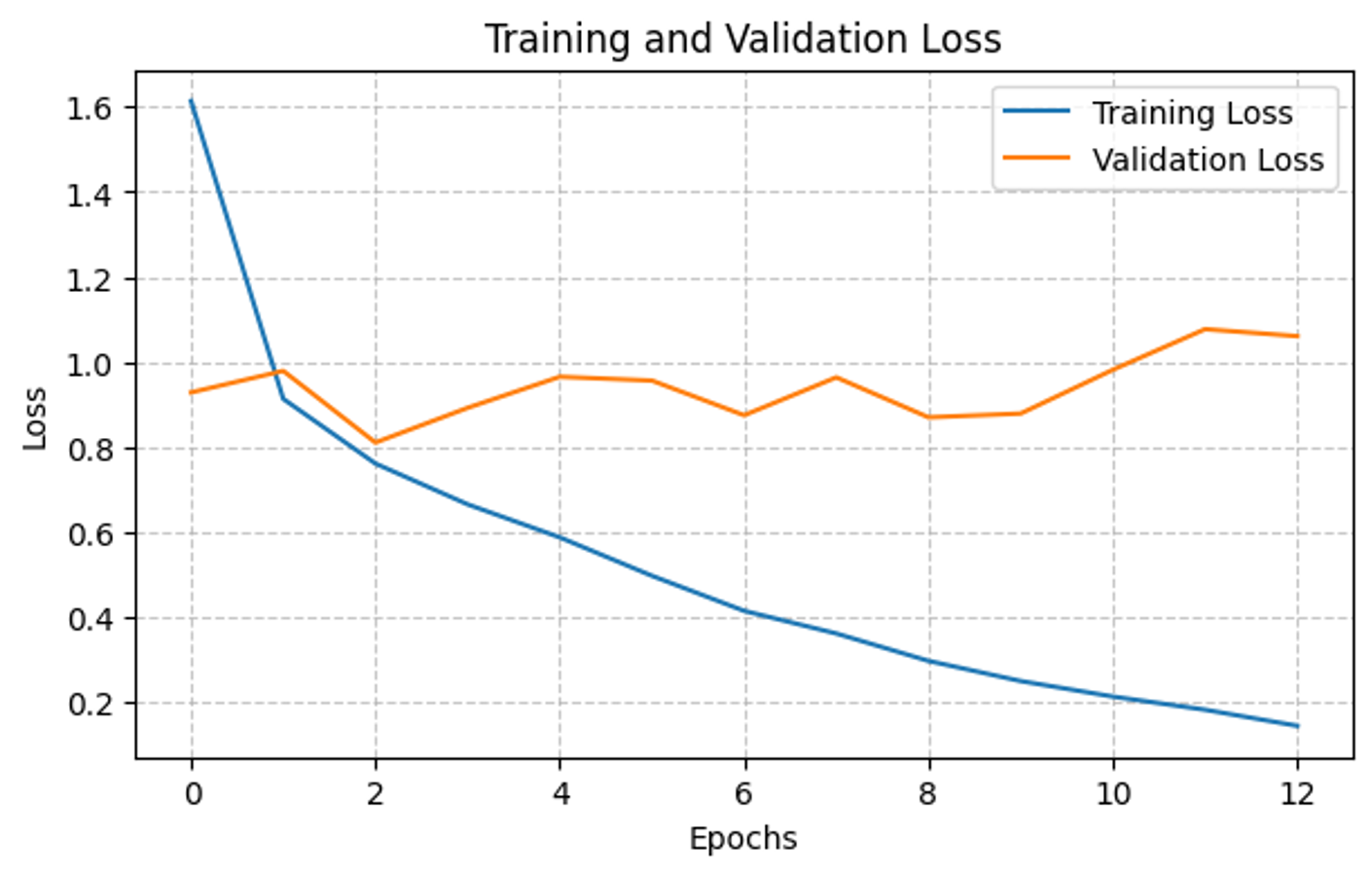}}
\centerline{\includegraphics[width=0.4\textwidth]{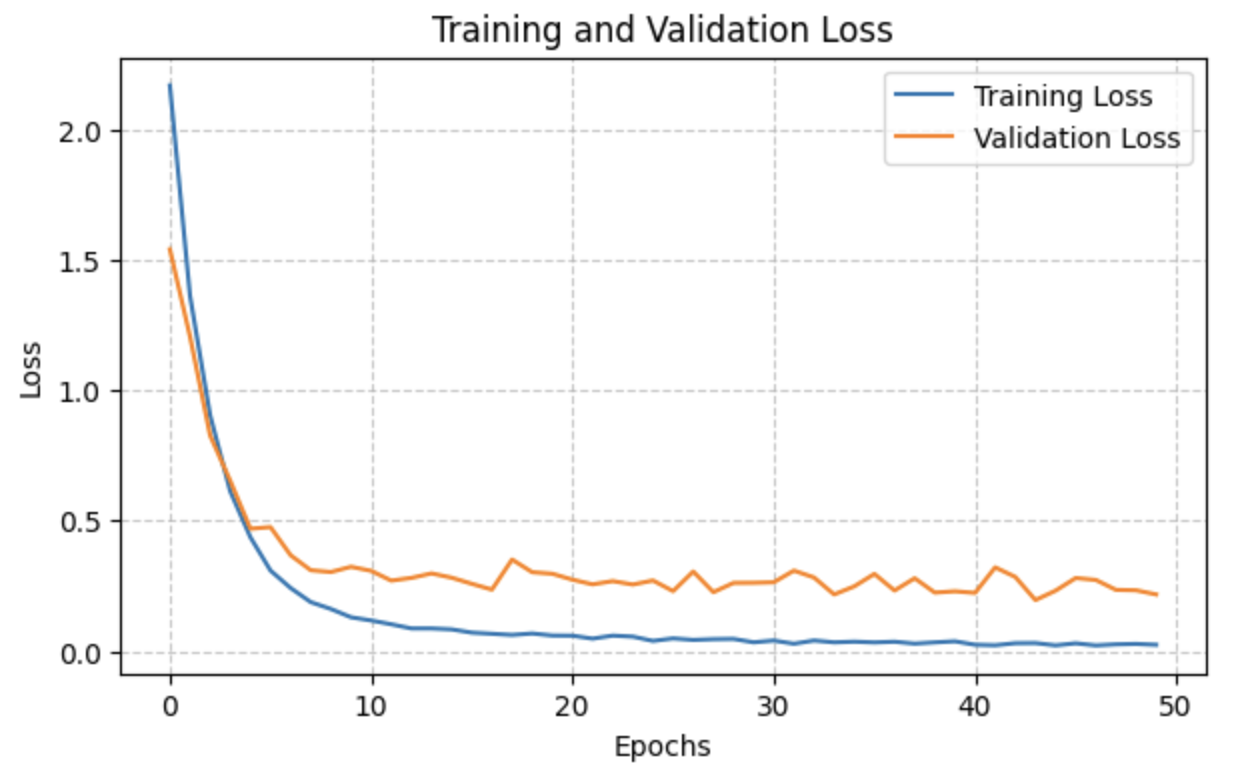}}
\caption{Loss comparison of InceptionV3 before and after optimization}
\label{fig}
\end{figure}

\begin{figure}[htbp]
\centerline{\includegraphics[width=0.4\textwidth]{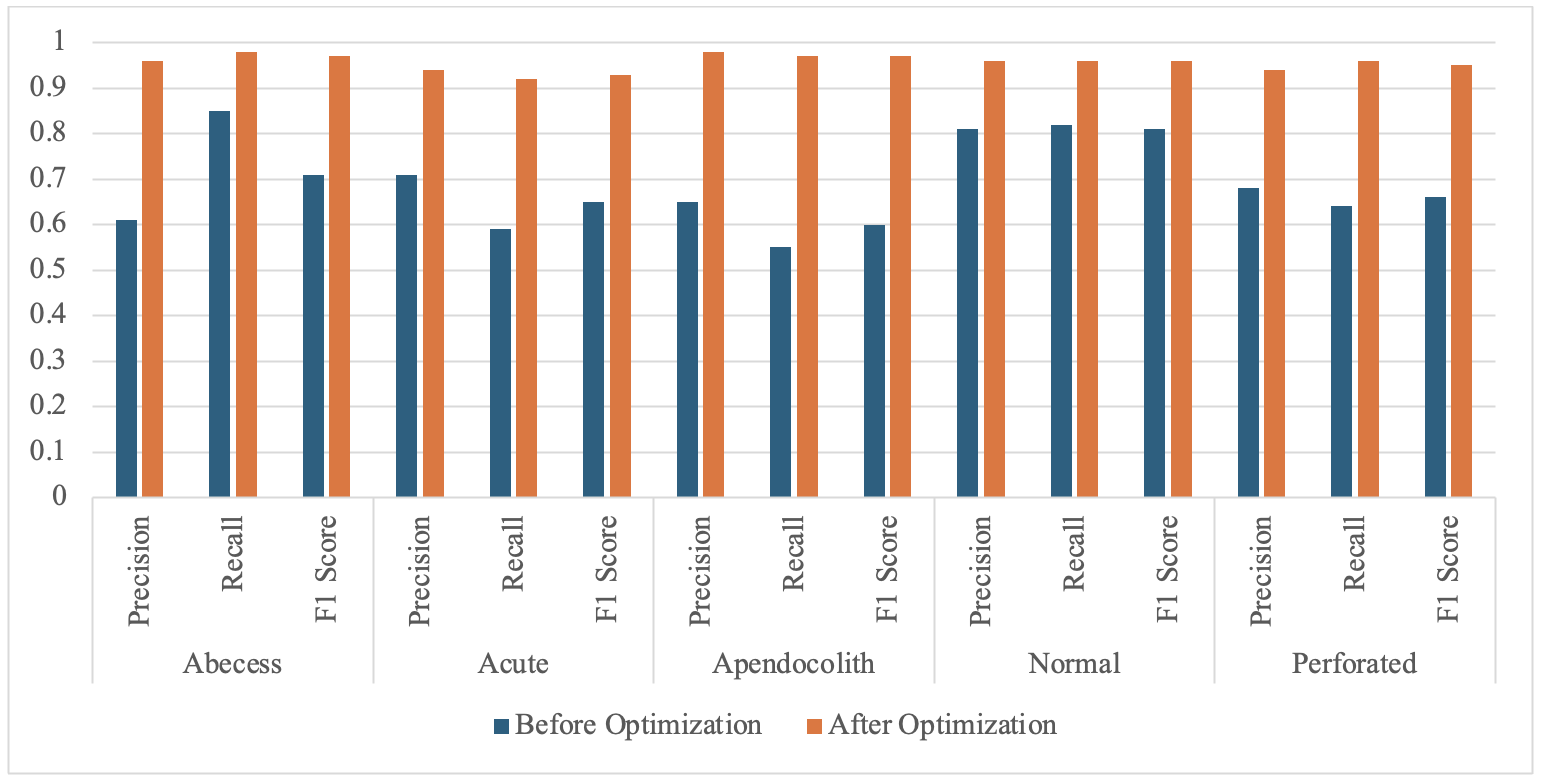}}
\caption{Comparison between before and after optimization for InceptionV3}
\label{fig}
\end{figure}

The remaining two classes showed similar improvements. For the perforated class the precision increased by 26\%, the recall increased by 31\%, and the F1 score increased by 30\%. For the normal class, the change was minimal, as the model's performance for this class was already the best among all classes. After enhancement, the precision, recall, and F1 score all improved by 14-15\%. The confusion matrix Fig. 9 for InceptionV3 after optimization highlights its improved accuracy. Only 7 abscess classes, 27 acute classes, 9 appendicolith classes, 14 normal classes and 14 perforated classes of ultrasound images were misclassified by the inception model. Overall, the model correctly classified 1,537 out of 1,608 ultrasound images.
\begin{figure}[htbp]
\centerline{\includegraphics[width=0.45\textwidth]{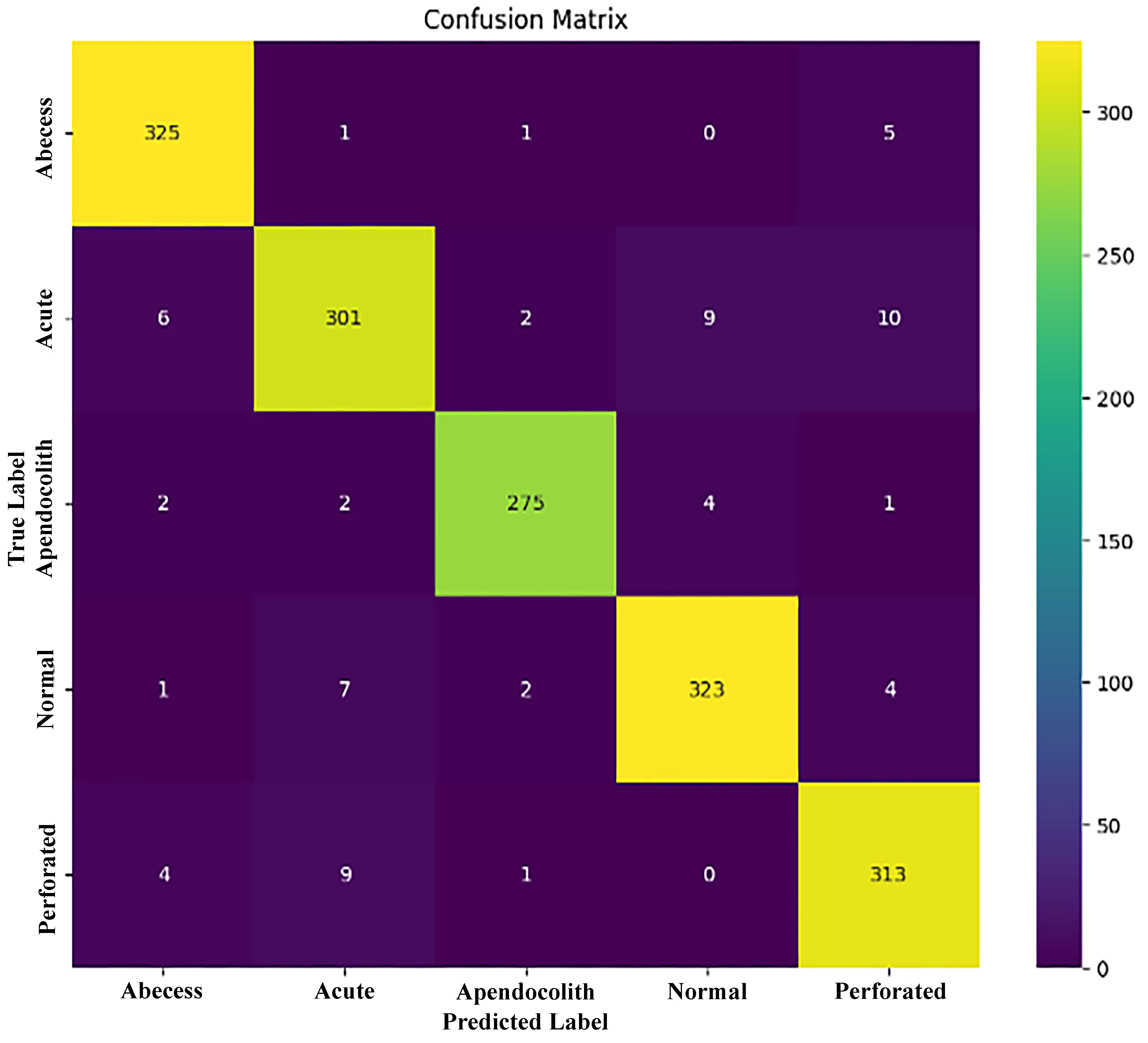}}
\caption{Confusion matrix for InceptionV3 after optimization}
\label{fig}
\end{figure}

\begin{figure}[htbp]
\centerline{\includegraphics[width=0.45\textwidth]{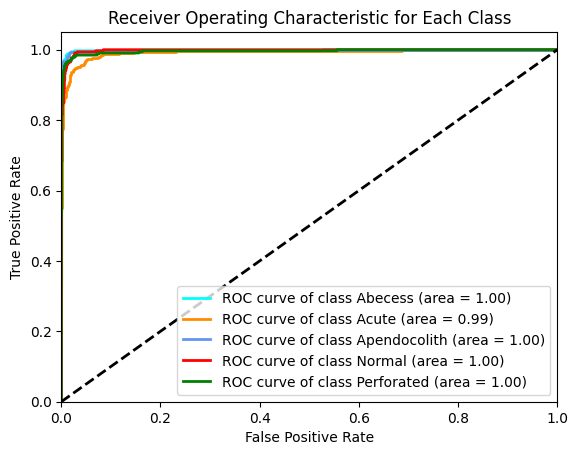}}
\caption{ROC curve for InceptionV3 after optimization}
\label{fig}
\end{figure}

The ROC curve Fig. 10 for our best-performing model, InceptionV3, is shown. The model achieved a perfect AUC score of 1.0 for four classes and a near-perfect 0.99 for the remaining class, demonstrating excellent performance across all classes. Model performance on the test dataset is shown in Fig. 11.
\begin{figure}[htbp]
\centerline{\includegraphics[width=0.5\textwidth]{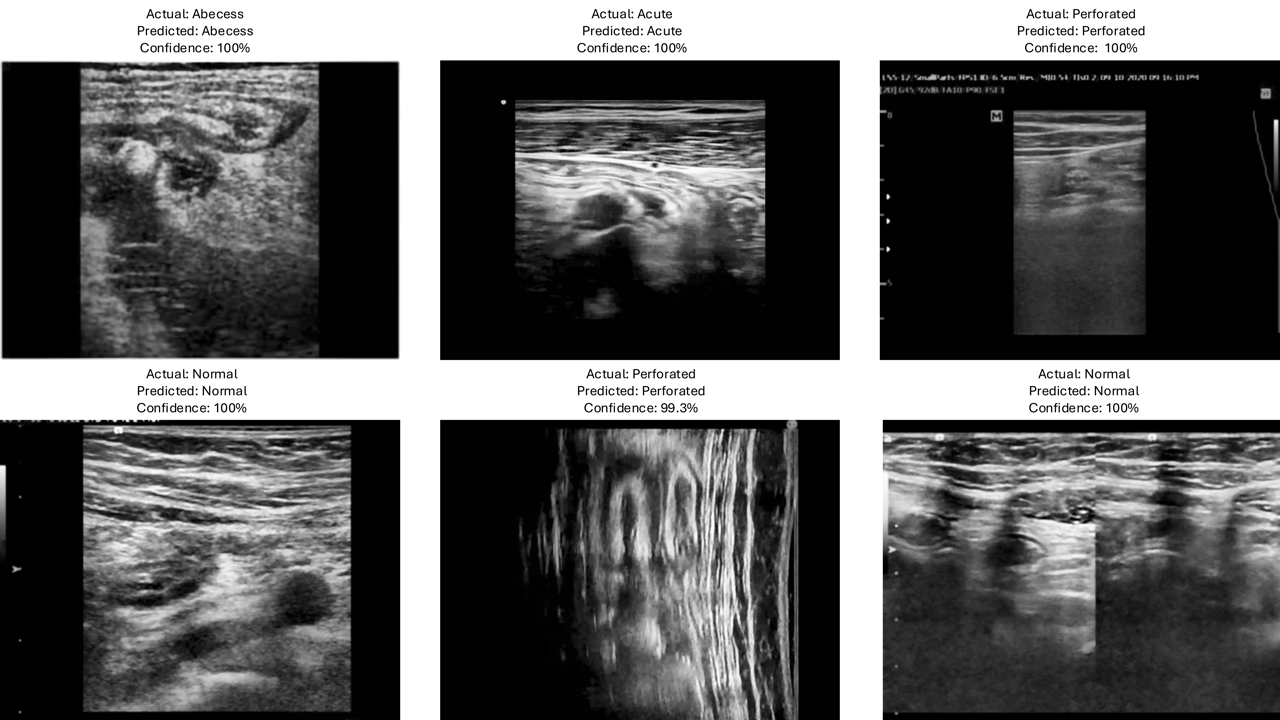}}
\caption{Model prediction on test data for each class}
\label{fig}
\end{figure}

Grad-CAM was integrated to visualize model predictions. The heatmap Fig. 12 highlights regions influencing classification decisions: red indicates the greatest influence, yellow moderate, and green to blue low to very low importance. These visualizations confirm that the areas visually distinguishing each type of complicated appendicitis are primarily responsible for the model's correct predictions.
\begin{figure}[htbp]
\centerline{\includegraphics[width=0.5\textwidth]{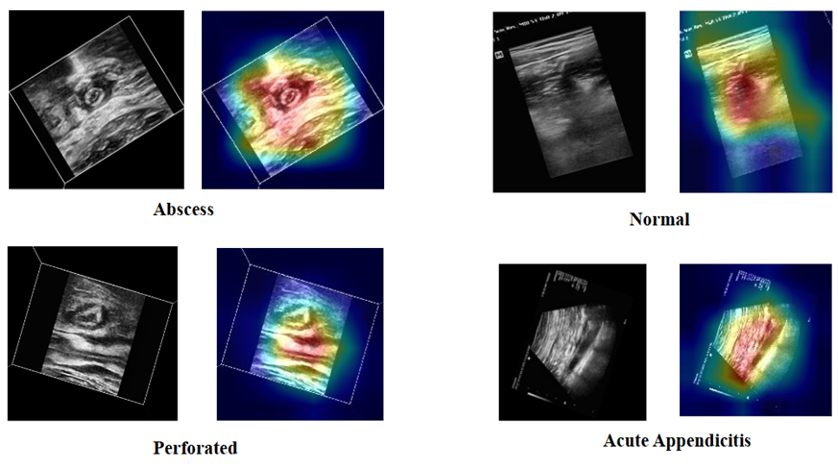}}
\caption{Grad-CAM heatmap for each type of complicated appendicitis for InceptionV3}
\label{fig}
\end{figure}



\section{Conclusion}
This study investigates different deep learning models to achieve the best results for detecting various types of appendicitis. Using default model setups on raw images yielded suboptimal results, with overfitting and significant disparities between training and testing accuracy. By leveraging multiple optimization techniques, substantial improvements were achieved, with InceptionV3 accuracy increasing from 69.21\% to 95.38\%. To improve interpretability, Grad-CAM was incorporated to provide visual details into the model’s decision-making process. The system has the potential to be integrated into healthcare, assisting physicians by highlighting areas of concern in ultrasound images and offering probable classification. The current study has several limitations: the dataset may not fully represent real-world variations, image sharpening can obscure details, heatmaps have not been verified by doctors, and the model has not been tested in diverse hospital settings. To address these, future research will focus on refining the system through clinical trials and pilot testing in collaboration with hospitals.



\begin{thebibliography}{00}

\bibitem{b1} A. Petroianu, “Acute appendicitis – propedeutics and diagnosis,” InTech eBooks, 2012.

\bibitem{b2} Y. Yang \textit{et al.}, “The global burden of appendicitis in 204 countries and territories from 1990 to 2019,” Clinical Epidemiology, vol. 14, pp. 1487–1499, 2022.

\bibitem{b4} M. M. Fischmann and F. Bellolio, “Acute appendicitis,” in Wiley, pp. 457–471, 2023.

\bibitem{b5} S. Malik, J. E. Harris, R. Vasdev, and A. T. Rezcallah, “Perforated appendicitis presenting as a soft tissue infection of the thigh: An example of successful non-operative management,” Cureus, 2023.

\bibitem{b6} L. P. H. Andersen, T. Pedersen, and J. Rosenberg, “Femoral abscess after acute appendicitis,” European Geriatric Medicine, 2012.

\bibitem{b7} S. P. Dhital \textit{et al.}, “Appendicoliths: The neglected stones,” Journal of Patan Academy of Health Sciences, 2017.

\bibitem{b8} J. Byun, S. Park, and S. M. Hwang, “Diagnostic algorithm based on machine learning to predict complicated appendicitis in children using CT, laboratory, and clinical features,” Diagnostics, vol. 13, no. 5, p. 923, 2023.

\bibitem{b9} W. A. Al, I. D. Yun, and K. J. Lee, “Reinforcement learning-based automatic diagnosis of acute appendicitis in abdominal CT,” arXiv preprint, 2019.

\bibitem{b12} M. M. Mijwil and K. Aggarwal, “A diagnostic testing for people with appendicitis using machine learning techniques,” Multimedia Tools and Applications, vol. 81, no. 5, pp. 7011–7023, 2022.

\bibitem{b10} P. Rajpurkar \textit{et al.}, “AppendiXNet: Deep learning for diagnosis of appendicitis from a small dataset of CT exams using video pretraining,” Scientific Reports, vol. 10, 2020.

\bibitem{b13} A. Pati \textit{et al.}, “Predicting pediatric appendicitis using ensemble learning techniques,” Procedia Computer Science, vol. 218, pp. 1166–1175, 2023.

\bibitem{b14} S. Akbulut \textit{et al.}, “Prediction of perforated and nonperforated acute appendicitis using machine learning-based explainable artificial intelligence,” Diagnostics, vol. 13, no. 6, p. 1173, 2023.

\bibitem{b15} R. Marcinkevičs \textit{et al.}, “Using machine learning to predict the diagnosis, management and severity of pediatric appendicitis,” Frontiers in Pediatrics, vol. 9, 2021.

\bibitem{b16} Ö. F. Akmeşe \textit{et al.}, “The use of machine learning approaches for the diagnosis of acute appendicitis,” Emergency Medicine International, 2020.

\bibitem{b11} J. J. Park \textit{et al.}, “Convolutional-neural-network-based diagnosis of appendicitis via CT scans in patients with acute abdominal pain,” Scientific Reports, vol. 10, 2020.

\bibitem{b17} P. Poortman \textit{et al.}, “Comparison of CT and sonography in the diagnosis of acute appendicitis: A blinded prospective study,” American Journal of Roentgenology, vol. 181, no. 5, pp. 1355–1359, 2003.

\bibitem{b18} M. Alelyani \textit{et al.}, “Evaluation of ultrasound accuracy in acute appendicitis diagnosis,” Applied Sciences, vol. 11, no. 6, p. 2682, 2021.

\bibitem{b19} A. R. Ö. Erdaş and Ç. B. Erdaş, “Computer-aided diagnosis of acute appendicitis in pediatric patients using deep learning models on abdominal ultrasound images,” in Proc. Int. Academic Conf. Global Education, Teaching and Learning, Dec. 2024.

\bibitem{b20} “abdoDetection Computer Vision Project,” Roboflow Universe, 2024.

\bibitem{b21} P. Baheti, “Train test validation split: How to \& best practices,” V7 Labs, 2021.

\bibitem{b3} M. E. Heilbrun \textit{et al.}, “Assessing the training costs and work of diagnostic radiology residents using key performance indicators,” Academic Radiology, vol. 27, no. 7, pp. 1025–1032, 2020.

\end{thebibliography}
\end{document}